\documentclass[11pt,a4paper]{article}
\usepackage[hyperref]{acl2020}
\usepackage{times}
\usepackage{url}
\usepackage{latexsym}
\usepackage{xcolor}
\usepackage[linesnumbered,ruled,vlined,noend]{algorithm2e}
\SetKwIF{If}{ElseIf}{Else}{if}{}{else if}{else}{end if}%
\SetKwFor{While}{while}{}{end while}%
\SetKwRepeat{Do}{do}{while}
\SetKw{KwGoTo}{go to}

\SetCommentSty{mycommfont}
\usepackage{microtype}
\usepackage{booktabs}
\usepackage{tabularx}
\usepackage{xcolor}
\usepackage{graphicx}
\usepackage{wrapfig}
\usepackage{amsmath}
\usepackage{dcolumn}
\usepackage{amssymb}
\usepackage{qtree}

\makeatletter
\def\blfootnote{\xdef\@thefnmark{}\@footnotetext}
\makeatother

\definecolor{sl}{rgb}{1.0, 0.75, 0.0}
\definecolor{iccjcc}{RGB}{172,229,133}
\definecolor{jcc}{RGB}{168,211,238}
\definecolor{icc}{RGB}{254,255,184}

\newcommand{\F}{$\textrm{F}_1$\xspace}
\newcommand{\ie}{{i.e.}\xspace}
\newcommand{\eg}{{e.g.}\xspace}
\newcommand{\ia}{{i.a.}\xspace}

\newcommand{\dsET}{\textit{ElectoralTweets}\xspace}
\newcommand{\dsGNE}{\textit{Good\-News\-Everyone}\xspace}
\newcommand{\dsREMAN}{\textit{REMAN}\xspace}
\newcommand{\dsECA}{\textit{Emotion\-Cause\-Analysis}\xspace}
\newcommand{\dsES}{\textit{Emotion\-Stimulus}\xspace}
\newcommand{\err}[1]{\includegraphics[page=#1, height=2.63009172mm]{errors}\xspace}
\usepackage{multirow}
\newcommand{\rtb}[1]{\rotatebox{90}{#1}}
\newcommand{\iob}[1]{\texttt{#1}}
\newcommand{\iO}{\iob{O}}
\newcommand{\iI}{\iob{I}}
\newcommand{\iB}{\iob{B}}
\renewcommand{\paragraph}[1]{\noindent\textbf{#1}}
\aclfinalcopy

\title{Token Sequence Labeling vs.\ Clause Classification for\\ English Emotion Stimulus Detection}

\author{Laura Oberl\"ander \and Roman Klinger\\
    Institut f{\"u}r Maschinelle Sprachverarbeitung, University of Stuttgart \\
    Pfaffenwaldring 5b, 70569 Stuttgart, Germany \\
   \texttt{\{laura.oberlaender,roman.klinger\}@ims.uni-stuttgart.de}
}
\date{}

\begin{document}
\maketitle
\begin{abstract}
  Emotion stimulus detection is the task of finding the cause of an
  emotion in a textual description, similar to target or aspect
  detection for sentiment analysis. Previous work approached this in
  three ways, namely (1) as text classification into an inventory of
  predefined possible stimuli (``Is the stimulus category \textit{A}
  or \textit{B}?''), (2) as sequence labeling of tokens (``Which
  tokens describe the stimulus?''), and (3) as clause classification
  (``Does this clause contain the emotion stimulus?''). So far,
  setting (3) has been evaluated broadly on Mandarin and (2) on
  English, but no comparison has been performed. Therefore, we analyze
  whether clause classification or token sequence labeling is better
  suited for emotion stimulus detection in English. We propose an
  integrated framework which enables us to evaluate the two different
  approaches comparably, implement models inspired by state-of-the-art
  approaches in Mandarin, and test them on four English data sets from
  different domains. Our results show that token sequence labeling is
  superior on three out of four datasets, in both clause-based and
  token sequence-based evaluation. The only case in which clause
  classification performs better is one data set with a high density
  of clause annotations. Our error analysis further confirms
  quantitatively and qualitatively that clauses are not the
  appropriate stimulus unit in English.
\end{abstract}

\blfootnote{
\hspace{-0.65cm}
This work is licensed under a Creative Commons Attribution 4.0 International License. License details: \url{http://creativecommons.org/licenses/by/4.0/}.}
\section{Introduction}
Research in emotion analysis from text focuses on classification, \ie, mapping
sentences or documents to emotion categories based on psychological theories
(\eg, \newcite{Ekman1992}, \newcite{Plutchik2001}). While this task answers the
question \emph{which} emotion is expressed in a text, it does not detect the
textual unit, which reveals \emph{why} the emotion has been developed. For
instance, in the example \emph{``Paul is angry because he lost his wallet.''}
it remains hidden that \emph{lost his wallet} is the reason for experiencing
the emotion of anger.
This stimulus, \eg, an event description, a person, a state of affairs, or an
object enables deeper insight, similar to targeted or aspect-based sentiment
analysis \cite[\ia{}]{Jakob2010a,Yang2013,Klinger2013,Pontiki2015,Pontiki2016}.
This situation is dissatisfying for (at least) two reasons. First, detecting
the emotions expressed in social media and their stimuli might play a role in
understanding why different social groups change their attitude towards
specific events and could help recognize specific issues in society. Second,
understanding the relationship between stimuli and emotions is also compelling 
from a psychological point of view, given that emotions are commonly considered
responses to relevant situations \cite{Scherer2005}.

Models which tackle the task of detecting the stimulus in a text have
seen three different problem formulations in the past: (1)
Classification into a predefined inventory of possible stimuli
\cite{Mohammad2014}, similarly to previous work in sentiment analysis
\cite{Ganu2009}, (2) classification of precalculated or annotated
clauses as containing a stimulus or not \cite[i.a.]{Gui2016}, and (3)
detecting the tokens that describe the stimulus, e.g., with IOB
labels \cite[i.a.]{Ghazi2015}.  We follow the two settings in which
the stimuli are not predefined categories (2+3, cf.\
Figure~\ref{fig:formulations}).

\begin{figure}
  \centering
    \centering\small
    \centering Clause-based Classification:\\[0.2em]
    \begin{tabular}{|cc|}
      \hline
      \texttt{No Stimulus}&\texttt{Stimulus}\\
      \enspace [ She's  pleased at ] & [ how  things  have turned out . ]\\
      \hline
    \end{tabular} \\[0.5em]
    \setlength{\tabcolsep}{3.2pt}
    \centering Token Sequence Labeling: \\[0.2em]
    \begin{tabular}{|cccccccccc|}
      \hline
      \iO & \iO & \iO & \iO & \iB & \iI & \iI & \iI & \iI & \iO \\
      She & 's & pleased & at & how & things & have & turned & out & . \\
      \hline
    \end{tabular}
    \caption{Different formulations for emotion stimulus detection.}
    \label{fig:formulations}
\end{figure}
 
 \begin{figure}
    \centering
    \includegraphics[scale=1.1]{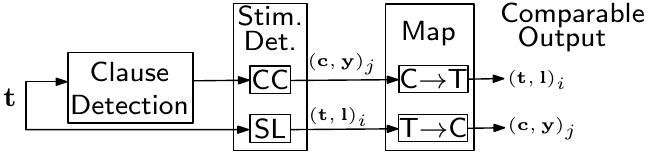}
    \caption{Framework for emotion stimulus detection. Tokens
      $\textbf{t}$ are split into clauses for clause class. Mapping
      ensures that both methods result in clause classifications
      $(\textbf{t},\textbf{l})_i$ and token sequences with labels
      $(\textbf{c},\textbf{y})_j$.}
    \label{fig:frame}
\end{figure}

These two settings have their advantages and disadvantages. The clause
classification setting is more coarse-grained and, therefore, more
likely to perform well than the token sequence labeling setting, but
it might miss the exact starting and endpoints of a stimulus span and
needs clause annotations or a syntactic parse with the risk of error
propagation. The token sequence labeling setting might be more
challenging, but has the potential to output more exactly which tokens
belong to the stimulus. Further, sequence labeling is a more standard
machine learning setting than a pipeline of clause detection and
classification.

These two different formulations are naturally evaluated in two
different ways and have not been compared before, to the best of our
knowledge. Therefore, it remains unclear which task formulation is
more appropriate for English.  Further, the most recent approaches
have been evaluated only on Mandarin Chinese, with the only exception
being the \dsECA dataset being considered by \newcite{fan2019}, but
not in comparison to token sequence labeling. No other English emotion
stimulus data sets have been tackled with clause classification
methods.
We hypothesize that clauses are not appropriate units for English, as
\newcite{Ghazi2015} already noted that: ``such granularity [is] too
large to be considered an emotion stimulus in English''. A similar
argument has been brought up during the development of semantic role
labeling methods: \newcite{Punyakanok2008} stated that ``argument[s]
may span over different parts of a sentence''.

Our contributions are as follows: (1) we develop an integrated
framework that represents different formulations for the emotion
stimulus detection task and evaluate these on four available English
datasets; (2) as part of this framework, we propose a clause detector
for English which is required to perform stimulus detection via clause
classification in a real-world setting; (3) show that token sequence
labeling is indeed the preferred approach for stimulus detection in
most available English datasets; (4) show in an error analysis that
this is mostly because clauses are not the appropriate unit for
stimuli in English. Finally, (5), we make our implementation and
annotations for both clauses and tokens available at
\url{http://www.ims.uni-stuttgart.de/data/emotion-stimulus-detection}.

The remainder of the paper is organized as follows. We first introduce
our integrated framework of stimulus detection which enables us to
evaluate clause classification and token sequence labeling in a
comparable manner (Section~\ref{sec:methods}). We then turn to the
experiments (Section~\ref{sec:experiments}) in which we analyze
results on four different English data sets. Section~\ref{sec:errors}
discusses typical errors in detail, which leads to a better
understanding of how stimuli are formulated in English.  We conclude
in Section~\ref{sec:conclusions}.

\section{An Integrated Framework for Stimulus Detection}
\label{sec:methods}
\vspace{-1mm}
The two approaches for open-domain stimulus detection, namely, clause
classification and token sequence labeling, have not been compared on English. We
propose an integrated framework (Figure~\ref{fig:frame}) which takes tokens
$\textbf{t}$ as input, splits this sequence into clauses and classifies them
(clause detection can be bypassed if manual annotations of clauses are
available). The token sequence labeling does not rely on clause annotations.
The output, either clauses $\textbf{c}$ with classifications $\textbf{y}$
($\textbf{y}\in\{\textrm{yes},\textrm{no}\}^n$) or tokens $\textbf{t}$ with
labels $\textbf{l}$ are then mapped to each other to enable a comparative
evaluation. We explain these steps in the following subsections.
\vspace{-1mm}

\subsection{Clause Extraction}

The clause classification methods rely on representing an instance as a
sequence of clauses. Clauses in English grammar are defined as the smallest
grammatical structures that contain a subject and a predicate, and can express
a complete proposition \cite{kroeger2005analyzing}. We show our algorithm to
detect clauses in Algorithm~\ref{fig:clausedetectionalg}.

\vspace{-1mm}
\begin{algorithm}[t]
\small
\RestyleAlgo{boxruled}
\SetAlgoLined
\DontPrintSemicolon
\LinesNumbered
\KwIn{text}
\KwOut{Clauses $\mathbf{c}$}
\SetKw{KwLoopIn}{in}
\SetKw{KwMyReturn}{return}
 $\mathbf{t}$ $\leftarrow$ tokenize(text)\;
 tree $\leftarrow$ parse($\mathbf{t}$)\tcp*{constituency parse}\label{alg:parse}
gaps $\leftarrow \{0, |\mathbf{t}|\}$ \tcp*{potential clause bounds} 
segments $\leftarrow \varnothing$ \tcp*{initial. set of segments}
\ForEach{node $n$ \KwLoopIn tree}{
  \If{{\rm label}(n) $\in$ S, SBAR, SBARQ, INV, SQ}{
    $\ell$ $\leftarrow$ first token leaf that $n$ governs\;
    $r$ $\leftarrow$ last token leaf that $n$ governs\;
    gaps = gaps $\cup\; \{$idx$_\ell$, idx$_r$ + 1$\}$\label{alg:node}
  }
}
\ForEach{adjacent pair $(i, j)$ \KwLoopIn sort(gaps)}{
    segments = segments $\cup{}\; \mathbf{t}[i:j]$
  }
\Repeat{convergence}{\label{alg:join}
   \ForEach{$s_i$ \KwLoopIn segments}{
        \If{$s_i \sim=$ {\tt /\^{}[\^{}A-za-z0-9]+\$/}}{
            $s_{i-1} = s_{i-1} \mathbin\Vert s_{i}$ \;
            segments = segments$\setminus s_{i}$
        }
        \If{$|s_i|\leq 3$}{
            $s_{i+1} = s_{i} \mathbin\Vert s_{i+1}$ \;
            segments = segments$\setminus s_{i}$

        }
   }
}
\KwMyReturn{segments}
\caption{Clause Extraction}
\label{fig:clausedetectionalg}
\end{algorithm}

\begin{figure}
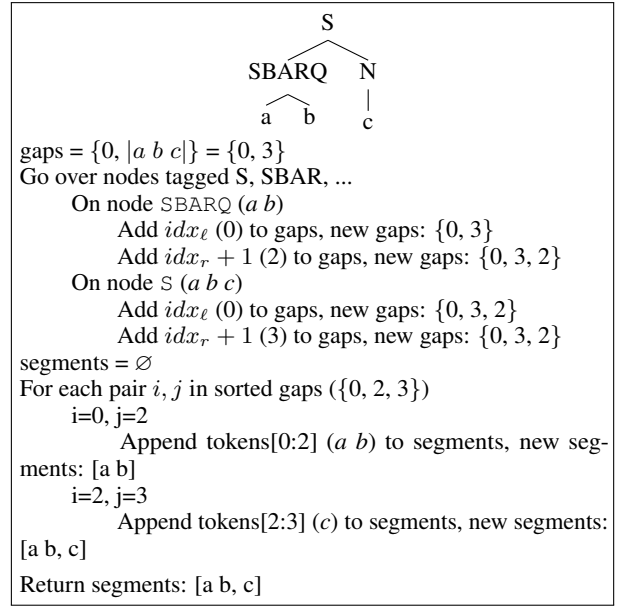

  \centering
  \fbox{\parbox{\linewidth}{%
\begingroup
\small
\Tree [.S [.SBARQ a b ] [.N c ] ]

gaps = \{0, $|a\; b\; c|$\} = \{0, 3\}

Go over nodes tagged S, SBAR, ...

\hspace*{6mm} On node \texttt{SBARQ} (\textit{a b})

\hspace*{12mm} Add $idx_\ell$ (0) to gaps, new gaps: \{0, 3\}

\hspace*{12mm} Add $idx_r + 1$ (2) to gaps, new gaps: \{0, 3, 2\}

\hspace*{6mm} On node \texttt{S} (\textit{a b c})

\hspace*{12mm} Add $idx_\ell$ (0) to gaps, new gaps: \{0, 3, 2\}

\hspace*{12mm} Add $idx_r + 1$ (3) to gaps, new gaps: \{0, 3, 2\}

segments = $\varnothing$

For each pair $i, j$ in sorted gaps (\{0, 2, 3\})

\hspace*{6mm} i=0, j=2

\hspace*{12mm} Append tokens[0:2] (\textit{a b}) to segments, new segments: [a b]

\hspace*{6mm} i=2, j=3

\hspace*{12mm} Append tokens[2:3] (\textit{c}) to segments, new segments: [a b, c]

Return segments: [a b, c]
\endgroup
}}
  \caption{Example for the application of Algorithm \ref{fig:clausedetectionalg}.}
  \label{fig:example}
\end{figure}

To mark the segments that would potentially approximate clauses, we rely on the
constituency parse tree of the token sequence (Line~\ref{alg:parse}). For that
reason, we use the Berkeley Neural Parser \cite{Kitaev2018}. As illustrated by
\newcite{feng2012} and \newcite{tafreshi2018} we also do that by segmenting the
constituency parse tree of the instance (Line~\ref{alg:node}) at the borders of
constituents labeled as clause-type \cite{Bies1995}. We then join the segments
until convergence heuristically based on punctuation (Line~\ref{alg:join}).
We illustrate the algorithm in the example in
Figure~\ref{fig:example}.

\subsection{Stimulus Detection}
Our goal is to compare sequence labeling and clause classification. To
attribute the performance of the model to the formulation of the task,
we keep the differences between the models at a minimum. We
therefore first discuss the model components and then
how we put them together.

Our models are composed of four layers.
As \textbf{Embedding Layer}, we use pretrained embeddings to embed each token
in the instance $s = t_1 \ldots t_n$ to obtain $\vec{e}_1, \ldots, \vec{e}_n$.
For the \textbf{Encoding Layer}, we use a bidirectional LSTM which
outputs a sequence of hidden states $\vec{h}_1, ..., \vec{h}_n$. In an
additional \textbf{Attention Layer}, each word or clause is
represented as the concatenation of its embedding and a weighted
average over other words or clauses in the instance:
$ \vec{u}_i = [\vec{h}_i ; \sum_{j=1}^{n} a_{i, j} \cdot \vec{h}_j
]$. The weights $a_{i, j}$ are calculated as the dot-product between
$\vec{h}_i$ and every other word, and by normalizing the scores using
softmax
$\vec{a}_i= \operatorname{softmax} (\vec{h}_i^T\cdot \vec{h}_j)$. We
concatenate all representations to obtain the final representation
vector $\vec{s}$.
The \textbf{Output Layer} is different for the two different task
formulations (sequence labeling vs.\ single softmax).
For the case of the single softmax, the input to the classifier is the
representation of the clause obtained on the previous layer and the classifier
output is defined as $\vec{o_i} = \operatorname{softmax}(W \cdot
\operatorname{ReLU}(\operatorname{Dropout}(h(\vec{s}))))$.
When labels are not predicted independently from each other but rather
in a sequential manner, we use a linear-chain conditional random field
\cite{Lafferty2001}. It takes the sequence of probability vectors from
the previous layer $\vec{u_1}, \vec{u_2}, \ldots $ and outputs a
sequence of labels $\vec{y_1}, \vec{y_2}, \ldots $. The score of the
labeled sequence is defined as the sum of the probabilities of
individual labels and the transition probabilities:
$s(y_{1:n})=\sum_{i=1}^n \vec{u}_i(y_i)+\sum_{i=2}^n T[y_{i-1}, y_i]$,
where the matrix $T$ that contains the transition probabilities
between one label and another (i.e., $T[i, j]$ represents the
probability that a token labeled $i$ is followed by a token labeled
$j$). At prediction time, the most likely sequence is chosen with the
Viterbi algorithm \cite{Viterbi1967}.

\begin{figure}
  \centering
   \includegraphics[scale=0.8]{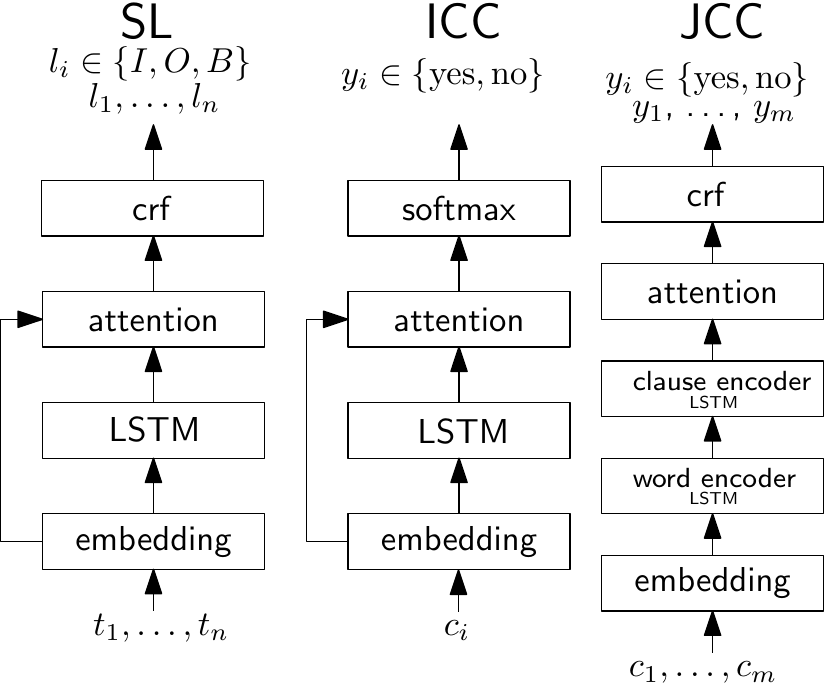}
   \caption{Comparable model architectures.}
   \label{fig:arch}
 \end{figure}

With these components, we can now put together the actual models which
we use for stimulus detection. We compare three different models, one
for token sequence labeling (SL) and two for clause classification
(CC). The model architectures are illustrated in
Figure~\ref{fig:arch}.

\paragraph{Token Sequence Labeling (SL).}
In this model, we formulate emotion stimulus detection as token
sequence labeling with the IOB alphabet \cite{iob}. As embeddings, we
use word-level GloVe embeddings \cite{glove}. The sequence-to-sequence
architecture comprises a bidirectional LSTM, an attention layer and
the CRF output layer.

\paragraph{Independent Clause Classification (ICC).}
This model, similarly proposed by \newcite{Cheng2017}, takes 
the clauses from the clause detector (or from annotated
data) and classifies them as containing the stimulus or not. The
model has a similar architecture to the one before, with the
exception of the final classifier, which is a single softmax to
output a single label. The training objective is to minimize the
cross-entropy loss. This model does not have access to clauses other
than the one it predicts for.

\paragraph{Joint Clause Classification (JCC).} In this model, the
neural architecture we employ is slightly different from before to
enable it to make a prediction for clauses in the context of all
clauses. It comprises multiple LSTM modules as word-level encoders,
one for each clause. The LSTM at the word-level encodes the tokens of
one clause into one representation. The next layer is a clause-level
encoder based on two bidirectional LSTMs, where the clause
representations are learned and updated by integrating the relations
between multiple clauses. After we obtain the final clause
representation for each clause, we perform sequence labeling with a
CRF \emph{on the clause level}. The training objective is to minimize
the negative log-likelihood loss across all clauses. This
implementation follows the architecture by
\newcite{Xia2019b}, with the change of the upper
layer, which is, in our case, an LSTM clause encoder and not a
transformer, to keep the architecture comparable across our different
formulations. Therefore, this is comparable to all other hierarchical
models proposed for the task \cite{Ding2019,Xu2019,Xia2019a}.

\subsection{Mapping between Task Formulations}
The last component of our integrated framework maps the different
representations of each formulation of emotion stimulus detection
between each other, namely clause classifications to token sequence labeling
and vice versa. We obtain clause classifications from token label
sequences ($T\rightarrow C$ in Figure~\ref{fig:frame}) by accepting
any clause that has at least one token being labeled as \texttt{B} or
\texttt{I} as a stimulus clause.  The other way around, clause classes
are mapped to tokens ($C\rightarrow T$) in such a way that the first
token of a stimulus clause is a \texttt{B} and all the
remaining tokens in the respective clause are \texttt{I}. Tokens from
clauses that do not correspond to a stimulus all receive \texttt{O}
labels.

\section{Experiments and Results}
\label{sec:experiments}
We now put the models to use to understand the differences between
sequence labeling and clause classification for English emotion
stimulus detection and the suitability of clauses as the unit of analysis.

\begin{table*}
  \centering\footnotesize
  \setlength{\tabcolsep}{6pt}
  \newcolumntype{d}[1]{D{.}{.}{#1}}
  \newcommand{\ccol}[1]{\multicolumn{1}{c}{#1}}
  \begin{tabular}{lcd{0}ccccd{0}d{0}ccccc} \toprule Data set & \ccol{Size} & \ccol{Stimuli} & \multicolumn{4}{c}{Tokens} & \multicolumn{4}{c}{Clauses}\\
    \cmidrule(lr){4-7}\cmidrule(l){8-11}
& & & $\mu$ & $\sigma$ & $\mu$S/I & $\mu$S/C & \ccol{Total} & \ccol{w. S} & $\mu$ I & $\mu$ w. all S/I \\
\cmidrule(r){1-1}\cmidrule(lr){2-2}\cmidrule(lr){3-3}\cmidrule(lr){4-4}\cmidrule(lr){5-5}\cmidrule(lr){6-6}\cmidrule(lr){7-7}\cmidrule(lr){8-8}\cmidrule(lr){9-9}\cmidrule(lr){10-10}\cmidrule(l){11-11}
\dsES & 2,414 & 820 & 7.29 & 5.20 & 0.12 & 0.11 & 5,818 & 1,117 & 2.41 & 0.05 \\ 
\dsET & 4,056 & 2,427 & 6.22 & 4.00 & 0.20 & 0.17 & 13,612 & 3,295 & 3.36 & 0.34 \\ 
\dsGNE & 5,000 & 4,798 & 7.27 & 3.67 & 0.55 & 0.50 & 9,190 & 6,301 & 1.84 & 0.52 \\ 
Emotion Cause Ana. & 2,655 & 2,580 & 8.48 & 5.20 & 0.20 & 0.10 & 19,473 & 2,897 & 7.33 & 0.33 \\ 
\bottomrule
  \end{tabular} \caption{Data sets available for the Emotion Stimulus Detection
    task in English. Size: number of annotated instances, Stimuli : number of
    instances with stimuli annotated; $\mu$, $\sigma$: mean/standard deviation of length
    of stimuli in tokens; $\mu$S/I: mean number of stimulus tokens per
    instance; $\mu$S/C: mean number of stimulus tokens per clause;
    Total: total number of clauses, w. S: number of clauses that
    contain a stimulus; $\mu$ I: average number of clauses
    per instance; $\mu$ w. all S/I: average number of clauses in which all tokens correspond to annotated stimuli.}
\label{tab:datasets}
\end{table*}

\subsection{Data Sets}
\label{sec:data}

We base our experiments on four data sets.\footnote{Corpora which we
  do not consider for our experiments are discussed in the related
  work section.} For each data set, we report the size, the number of
stimulus annotations and statistics for tokens and clauses in
Table~\ref{tab:datasets}.

\paragraph{\dsES.} This data set proposed by \newcite{Ghazi2015} is constructed
based on FrameNet's \emph{emotion-directed}
frame.\footnote{\url{https://framenet2.icsi.berkeley.edu/fnReports/data/frameIndex.xml?frame=Emotion_directed}}
The authors used FrameNet’s annotated data for 173 emotion lexical units,
grouped the lexical units into seven basic emotions using their synonyms and
built a dataset manually annotated with both the emotion stimulus and the
emotion. The corpus consists of 820 sentences with annotations of emotion
categories and stimuli. The rest of 1,594 sentences only contain an emotion
label. For this dataset, we see the lowest average number of clauses for which
all tokens correspond to a stimulus ($\mu$ w. all S/I in
Table~\ref{tab:datasets}). This result shows that the stimuli annotations
rarely align with the clause boundaries.

\paragraph{\dsET.} Frame Semantics also inspires a dataset of social media
posts \cite{Mohammad2014}. The corpus consists of 4,056 tweets of which 2,427
contain emotion stimulus annotations on the token level. The annotation was
performed via crowdsourcing. The tweets are the shortest instance type in
length and have a higher average of clauses per instance than the \dsGNE or the
\dsES datasets. They also show the same mean of stimulus tokens per instance as
\dsECA with a slightly higher mean for the number of clauses in which all
tokens correspond to stimulus annotations.

\paragraph{\dsGNE.} The data set by \newcite{Bostan2020} consists of
news headlines. From a total of 5000 instances, 4,798 contain a
stimulus. The headlines have the shortest stimuli in token
count. Similar to the \dsET, they also have a high average
stimulus token density in clauses. This set has the lowest mean number
of clauses per instance ($\mu$ I in Table~\ref{tab:datasets}).

\paragraph{\dsECA} \cite{Gao2017} comparably annotate English and
Mandarin texts on the clause level and the token level. In our work,
we use the English subset, which is the only English corpus annotated
for stimuli both at the clause level and at the token level. This
dataset has the fewest instances without stimuli among all the
others. It also has the longest instances and stimuli. The mean of
stimuli tokens annotated per clause is comparable to \dsES despite
having a higher mean of stimuli tokens per instance. In the upcoming
experiments, we use the clause annotations and not automatically
recognized clauses with Algorithm~\ref{fig:clausedetectionalg} as
input to our framework.

\begin{table*}
  \centering
  \footnotesize
  \begin{tabular}{l ccc ccc ccc c}
    \toprule
   && \multicolumn{6}{c}{Intrinsic} & \multicolumn{3}{c}{Extrinsic}  \\
   \cmidrule(rl){3-8}\cmidrule(rl){9-11}
    & IAA  & \multicolumn{3}{c}{Stimuli vs. Anno. Clauses} &  \multicolumn{3}{c}{Extra. vs. Anno. Clauses} & \multicolumn{3}{c}{Stimuli vs. Extra. Clauses}  \\
   \cmidrule(rl){2-2}\cmidrule(rl){3-5}\cmidrule(rl){6-8}\cmidrule(rl){9-11}
    Dataset & $\kappa$ & Exact & Left & Right & Precision  & Recall & F1 &  Exact & Left & Right  \\
    \cmidrule(r){1-1}\cmidrule(rl){2-2}\cmidrule(rl){3-3}\cmidrule(rl){4-4}\cmidrule(rl){5-5}\cmidrule(rl){6-6}\cmidrule(rl){7-7}\cmidrule(rl){8-8}\cmidrule(rl){9-9}\cmidrule(rl){10-10}\cmidrule(lr){11-11}
    \dsECA   & 0.60 & 0.60 & 0.35 & 0.86 & 0.77  & 0.75  & 0.76  & 0.59  & 0.36 & 0.84 \\ 
    \dsGNE  & 0.77 & 0.62 & 0.29 & 0.90 & 0.87 & 0.76 & 0.80 & 0.61 & 0.27 & 0.89 \\  
    \dsES    & 0.59 & 0.47 & 0.83 & 0.11 & 0.86 & 0.72 & 0.76 & 0.17  & 0.26 & 0.07   \\
    \dsET    & 0.63 & 0.56 & 0.39 & 0.63 & 0.82 & 0.78 & 0.80 &  0.54 & 0.43 & 0.60 \\
    \bottomrule
\end{tabular}
\caption{Evaluation of Clause Detection. Note that for \dsECA, the
  clauses stem from the annotation provided in the original data and
  not from our automatic detection method.}
\label{tab:clause-extraction}
\end{table*}

\subsection{Clause Identification Evaluation}
Before turning to the actual evaluation of the emotion stimulus detection
methods, we evaluate the quality of the automatic clause detection. For an
intrinsic evaluation, we annotate 50 instances from each test corpus in each
data set with two annotators trained on the clause extraction task in two
iterations. The two annotators are graduate students and have different
scientific backgrounds: computational linguistics (A1) and computer science
with a specialization in computer vision (A2). Each student annotated 50
instances of each dataset from the datasets we use in the same order. As an
environment for the annotation process, we used a simple spreadsheet application. We did
this small annotation experiment as an inner check for our understanding of the
clause extraction task. None of the annotators is a native English speaker; A1
is a native speaker of a Romance language, and A2 a German speaker. The
inter-annotator agreement is shown in Table~\ref{tab:clause-extraction}. We
achieve an acceptable average agreement of $\kappa$=.65.

We now turn to the question if annotated clauses (as an upper bound to
an automatic system) align well with annotated stimuli
(\textbf{Stimuli vs.\ Anno.\ Clauses} in
Table~\ref{tab:clause-extraction}). The evaluation is based on recall
(i.e., measuring for how many stimuli a clause exists), either for the
whole stimulus (exact), or for the left or the right boundary. We see
that except for the corpus \dsES, the right boundaries match better
than the left.

Turning to extracted clauses instead of annotated ones
(\textbf{Extra.\ vs.\ Anno.\ Clauses}) we first evaluate the automatic
extraction algorithm. We obtain \F values between 0.76\% and 0.80\%,
which we consider acceptable though they also show that error
propagation could occur.

For the actual extrinsic evaluation, if clause boundaries are correctly found
for annotated stimuli (\textbf{Stimuli vs.\ Extra. Clauses}), we see that the
results are only slightly lower than for the gold annotations, except for
\dsES. Therefore, we do not expect to see error propagation due to an imperfect
extraction algorithm for most data sets.

These results suggest that clauses are not an appropriate unit for
stimuli in English. Still, we do not know yet if the clause detection
task's simplicity outweighs these disadvantages in contrast to token
sequence labeling. We turn to answer this in the following.

\begin{figure*}[t]
\centering
\includegraphics[width=.92\linewidth]{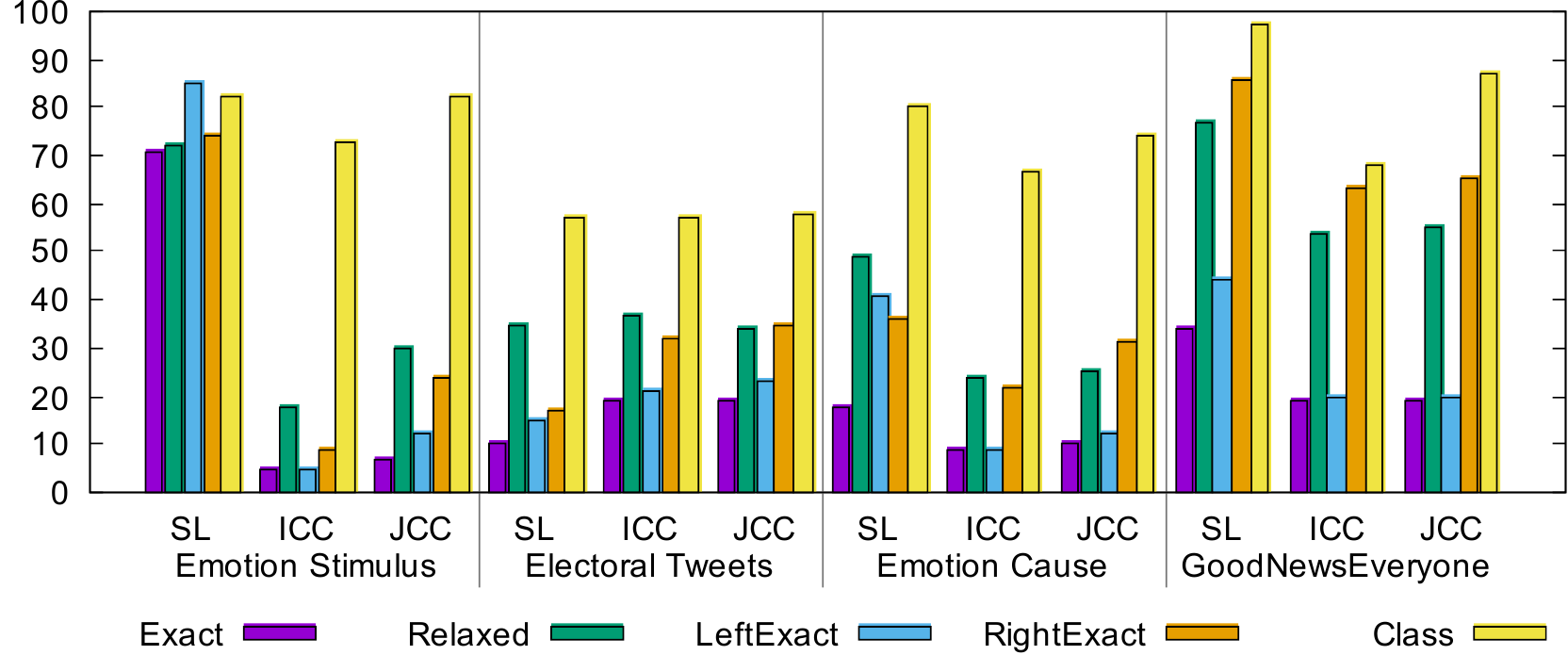}
\caption{Results of the three different models across four different datasets}
\label{fig:results}
\end{figure*}

\subsection{Stimulus Detection Evaluation}

\subsubsection{Evaluation Procedure}
We evaluate the quality of all models with five different
measures. Motivated by the formulation of \emph{clause
  classification}, we (1) evaluate the prediction on the clause level
with precision, recall, and \F. For the \emph{sequence labeling}
evaluation, we use four variations.  (2)~\emph{Exact}, where we
consider a consecutive token sequence to be correct if a gold
annotation exists that exactly matches, (3)~\emph{Relaxed}, where an
overlap of one token with a gold annotation is sufficient,
(4)~\emph{Left-Exact} and (5)~\emph{Right-Exact}, where at least the
most left/right token in the prediction needs to have a gold-annotated
counterpart.

One might argue that sequence labeling evaluation is unfair for the
clause classification, as it is more fine-grained than the actual
prediction method. However, for transparency across methods and
analysis of advantages and disadvantages of the different methods, we
use this approach in addition to clause classification evaluation.

We split the data for each set randomly into three sets: $80 \%$
train, $10 \%$ dev, and $10 \%$ test. We use dropout with a
probability of 0.5, train with Adam \cite{KingmaB14} with a base
learning rate of 0.003, and a batch size of 10. At test time, we
select the model with the best validation accuracy after 50 epochs
with a patience of 10 epochs. All models use embedding sizes of 300
and hidden state sizes of 100 \cite{glove}. We do not tune
hyperparameters for any of the architectures and implement all models
with the AllenNLP library \cite{allennlp}.

\begin{table*}
\setlength{\tabcolsep}{4pt}
\renewcommand{\arraystretch}{0.95}
\centering\footnotesize
\begin{tabular}{llrrr rrr rrr rrr r}
\toprule
& & \multicolumn{4}{c}{SL} & \multicolumn{4}{c}{ICC} & \multicolumn{4}{c}{JCC}  &  \\
\cmidrule(lr){3-6} \cmidrule(lr){7-10} \cmidrule(lr){11-14}
$\textrm{Ann.}\over\textrm{Pred.}$& Error types & ET & GNE & ES & ECA & ET & GNE & ES & ECA & ET & GNE & ES & ECA & Sum \\
\cmidrule(r){1-1}\cmidrule(r){2-2}\cmidrule(lr){3-6}\cmidrule(lr){7-10}\cmidrule(lr){11-14}\cmidrule(l){15-15}
\err{1}  & Early stop & 0  & 4 & 1 & 3 & 0 & 6 & 2 & 7 & 0 & 5 & 1 & 4  & 33\\
\err{2}  & Late stop &  11 & 9 & 10 & 8 & 19 & 30 & 7 & 25 & 17 & 31 & 13 & 22 & 202 \\
\cmidrule(r){1-1}\cmidrule(r){2-2}\cmidrule(lr){3-6}\cmidrule(lr){7-10}\cmidrule(lr){11-14}\cmidrule(l){15-15}
\err{3}  & Early start \& stop & 0 & 3 & 0 & 1 & 9 & 11 & 5 & 1 & 6 & 10 & 3 & 2 & 51 \\
\err{4}  & Early start &  152 & 16 & 0 & 6  & 192 & 73 & 9 & 164 & 220 & 58 & 3 & 159 & 1052 \\
\cmidrule(r){1-1}\cmidrule(r){2-2}\cmidrule(lr){3-6}\cmidrule(lr){7-10}\cmidrule(lr){11-14}\cmidrule(l){15-15}
\err{7}  & Late start & 28 & 3 & 0 & 1 & 3 & 8 & 1 & 0 & 2 & 7 & 1 & 0 & 54 \\
\err{8}  & Late start \& stop & 2  & 1 & 0 & 0  & 0 & 2 & 0 & 1 & 0 & 1 & 0 & 1 & 8 \\
\cmidrule(r){1-1}\cmidrule(r){2-2}\cmidrule(lr){3-6}\cmidrule(lr){7-10}\cmidrule(lr){11-14}\cmidrule(l){15-15}
\err{6}  & Contained &  0 & 0 & 0 & 0 & 0 & 0 & 0 & 1 & 0 & 0 & 0 & 2 & 3 \\
\err{10}  & Multiple & 143 & 189 & 11 & 260 &  47 & 24 & 9 & 11 & 37 & 34 & 4 & 0 & 769 \\
\err{5}  & Surrounded & 9 & 10 & 0 & 5 & 19 & 31 & 28 & 43  & 22 & 33 & 26 & 28 & 254 \\
\cmidrule(r){1-1}\cmidrule(r){2-2}\cmidrule(lr){3-6}\cmidrule(lr){7-10}\cmidrule(lr){11-14}\cmidrule(l){15-15}
\err{11}  & False Negative & 231 & 160 & 59 & 228 & 126 & 112 & 10 & 97 & 85 & 50 & 1 & 81 & 1240 \\
\cmidrule(r){1-1}\cmidrule(r){2-2}\cmidrule(lr){3-6}\cmidrule(lr){7-10}\cmidrule(lr){11-14}\cmidrule(l){15-15}
\err{9}/\err{12}  & False Positives  & 10 & 18  & 2 & 14 & 78 & 92 & 11 & 38 & 60 & 73 & 4 & 26  & 426 \\
\cmidrule(r){1-1}\cmidrule(r){2-2}\cmidrule(lr){3-6}\cmidrule(lr){7-10}\cmidrule(lr){11-14}\cmidrule(l){15-15}
& All & 586 & 413 &  83 &  526  & 493 & 389 & 82 & 388 & 449 & 302 & 56 &  325 & 4092  \\
\bottomrule
\end{tabular}
\caption{Counts for each error type for each model across all
  data sets.}
\label{tab:counts}
\end{table*}

\subsubsection{Results}

We now study the performance of the different models on the English data sets.
Figure~\ref{fig:results} summarizes the results. (Precision and recall
values are available in Table~\ref{tab:exp1} in Appendices.)

\paragraph{Which of the modeling approaches performs best on English data?}
If we only compare the absolute numbers in \F, we see that the clause
classification evaluation (Class) shows the highest result across all
models and data set. The only exception is the \dsES data, in which
the Left-Exact evaluation is slightly higher. When we rely on this
evaluation score, we see that the token sequence labeling method shows
a superior result to the classification methods in two data sets,
namely \dsGNE and \dsECA. On \dsET and \dsES, the results are
\textit{en par} across all methods with this evaluation measure. We
find this surprising to some degree, as this evaluation is more
natural for the classification tasks (ICC and JCC) than for sequence
labeling (SL), which requires the mapping step.

As this suggests that clauses are not the appropriate unit, it is
worth comparing these results with the Exact evaluation measure, which
evaluates on the token-sequence level. We observe that token sequence
labeling outperforms both clause classification methods on three of
the four data sets, with \dsET being the only exception with the
shortest textual instances and the highest number of clauses in which
all tokens correspond to stimulus annotation (see
Table~\ref{tab:datasets}). Therefore, we conclude that token sequence
labeling is superior to clause classification on (most of our) English
data sets.

\paragraph{Do clause classification models perform better on the left or the
right side of the stimulus clause?}
Given the evaluation of the clause detection, we expect the right boundary to
be better found for \dsGNE and \dsECA and the left boundary for \dsES.
Surprisingly, this is not entirely true -- the right boundary is found with
higher \F on all data sets, not only on those where the clauses are better
aligned with the stimulus' right boundary. Nevertheless, the effect is more
reliable for \dsGNE, as expected.

\paragraph{Does token sequence labeling perform better on the left or the right side
of the stimulus clause?}
We can ask this similar question for token sequence labeling, though it might be
harder to motivate than in the classification setting. Non-surprisingly, such a
clear pattern cannot be observed. For \dsET and \dsECA, the difference between
the left and right match is minimal. For \dsGNE, it can be observed to a lesser
extent than for the classification approaches, and for \dsES, the left boundary
is better found than the right boundary. It seems that for the longer sequences
in \dsES and \dsECA, the beginning of the stimulus span is easier to find than
for shorter sequences.

\paragraph{Is joint prediction of clause labels beneficial?}
This hypothesis can be confirmed; however, the differences are of a different
magnitude depending on the data set. For \dsGNE, the effect is more substantial
than for the other corpora. \dsET shows the smallest difference.

\begin{figure*}
\centering
\footnotesize
\renewcommand{\arraystretch}{0.86}
\setlength{\tabcolsep}{5pt}
\setlength{\fboxsep}{1pt}
\begin{tabular}{llll}
\toprule
Err  & Example & Model & Data set \\
\midrule
\err{1} & \textbf{\colorbox{lightgray}{Steve talked to me a lot \big| about being abandoned} \big| and the pain} \big| that caused.   & JCC & ECA\\
\err{2} & No what I told about \textbf{\colorbox{lightgray}{the way \big| they treated you} and me} \big| made him angry. & SL & ECA\\
\err{2}  &   \colorbox{lightgray}{\textbf{Fuck} Mitt Romney \big| and Fuck Barack Obama} \big| ... God got me !!!!!  & ICC & ET \\
\err{3}  & \colorbox{lightgray}{Maurice Mitchell \textbf{wants}} \big| \textbf{you to do more than vote}.  & ICC & GNE \\
\err{3}  & And he started \big| to despair \colorbox{lightgray}{that \textbf{his exploration was going}}  \big| \textbf{to be entirely unsuccessful ...} & ICC & ECA \\
\err{4}  & \colorbox{lightgray}{Deeply ashamed \textbf{of my wayward notions}}, \big| I tried my best to contradict myself. & ICC & ES\\
\err{5}  & Anyone else find it weird \big| I get excited about \colorbox{lightgray}{stuff like the \textbf{RNC} tonight} ?! \big| \# polisciprobs  & SL & ET\\
\err{6}  & \textbf{Doesn't \colorbox{lightgray}{he do it} well} \big| said the girl following with admiring eyes, \big| every movement of him. & JCC & ECA\\
\err{7}  &  If he feared \big| that \textbf{some \colorbox{lightgray}{terrible secret might evaporate from them}}, \big| it was a mania with him. & SL & ECA \\
\err{8}  & I was furious \big| \textbf{because} \colorbox{lightgray}{\textbf{the Mac XL wasn't real} said Hoffman}. & SL & ECA \\
\err{9}  &  With such obvious delight
\textbf{in food}, it 's hard  \big| \colorbox{lightgray}{to see how Blanc remains so slim. } & SL & ES\\
\err{10}  & \textbf{Triad Thugs Use Clubs \colorbox{lightgray}{to Punish Hong Kong} ’ s \colorbox{lightgray}{Protesters}.}& JCC & GNE\\
\err{11}  & \textbf{I'm glad to see you} \big| so happy Lupin  &  ICC & ES\\
\bottomrule
\end{tabular}
\caption{Examples for error types for different models and data
sets. Extracted clauses are separate by \big|.}
\label{fig:error_types}
\end{figure*}

\section{Error Analysis}
\label{sec:errors}
In the following, we analyze the error types made by the different models on
all data sets and investigate in which ways SL improves over the ICC and
JCC models. We hypothesize that the higher
flexibility of token-based sequence labeling leads to different types of errors than
the clause-based classification models.

For quantitative analysis, we define different error types,
illustrated in Table~\ref{tab:counts} with different symbols as
abbreviations. The top bar illustrates the gold span, while the bottom
corresponds to the predicted span.  The error types illustrated with
symbols \err{9} and \err{12} correspond to false positives; \err{11}
are false negatives. All other error types correspond to either both
false positive and false negative in a strict evaluation setting or
true positives in one of the relaxed evaluation settings.

\paragraph{Do ICC and JCC particularly miss starting or end points of the
stimulus annotation?} We see in Table~\ref{tab:counts} that for \textit{Late
stop} \err{2}, CC models make considerably more mistakes across all datasets.
ICC does so on ET and ECA, while JCC makes more mistakes on GNE and ES. For
data sets in which stimulus annotations end with a clause, errors of this type
are less likely. These results are more prominent for \textit{Early start \&
stop} \err{3}.

\paragraph{Do all methods have similar issues with finding the whole
consecutive stimulus?} We see this in the error type \textit{Multiple}
\err{10}. When the CC models make this mistake, it can be attributed to the
automatic fine-grained clause extraction, which can cause a small clause within
a gold span to become a false negative.
However, we see that SL shows higher numbers of this issue than CC. This
result is also reflected in the surprisingly low number of \textit{Contained}
(\err{6}) -- if the prediction is completely inside a gold annotation, the gold
annotation tends to be long, and this increases the chance that it is (wrongly)
split into multiple predictions.

\paragraph{How do the error types differ across \textit{models}?} The
\textit{Early Start (\& Stop)} and \textit{Surrounded} (\err{3},
\err{4}, \err{5}) counts show differences across the different types
of models.  Presumably, the clause classification models do have
difficulties in finding the left boundary, and they are more prone to
``start early'' than the token sequence labeling models. This might be
due to gold spans starting in the middle of a clause which is
predicted to contain the stimulus.

\paragraph{How do the error types differ across \textit{data sets}?}
The results and error types differ across data sets (see particularly
\err{10}, \err{4}, \err{3}). This points out what we have seen in the
evaluation already: The structure of a stimulus depends on the domain
and annotation. The least challenging data set is \dsES with the
lowest numbers of errors across all models. This result is caused by
most sentences having similar syntactic trees, all stimuli are
explicit and mostly introduced in a similar way.

For qualitative analyses, Figure~\ref{fig:error_types} shows one example of
each type of error described above. In the first example, the JCC model does
not learn to include the second part of the coordination -- ``and the pain''.
In the second example, similarly, the SL model misses the right part of the
coordination. For most cases of independent clauses that we inspect, we see a
common pattern for both types of models, which is that the prediction stops
while encountering coordinating conjunctions. In the sixth example, the
prediction span includes the emotion cue. This issue could be solved by doing
sequence labeling instead or by informing the model of the presence of other
semantic roles. These examples raise the following question: would improved
clause segmentation lead to improvements for the clause-classification models
across all data sets?

\section{Related Work} \label{sec:related}
The task of detecting the stimulus of an expressed emotion in text
received relatively little attention.

Next to the corpora we mentioned so far, the \textit{\dsREMAN} corpus
\cite{Kim2018} consists of English excerpts from literature, sampled
from Project Gutenberg. The authors consider triples of sentences as a
trade-of between longer passages and sentences. Further,
\newcite{Neviarouskaya2013} annotated English sentences on the token
level.

Besides English and Mandarin, \newcite{Russo2011} developed a method
for the identification of Italian sentences that contain an emotion
cause phrase.  \newcite{Yada2017} annotate Japanese sentences on
newspaper articles, web news articles, and Q\&A sites.
Table~\ref{tab:comparison} in Appendices shows which corpora and methods have
been used and compared in previous work for the available English and Chinese
sets. We see that the methods applied on the Chinese sets are not evaluated on
the English sets.

\newcite{Lee2010emotion} firstly investigated the interactions between
emotions and the corresponding stimuli from a linguistic
perspective. They publish a list of linguistic cues that help in
identifying emotion stimuli and develop a rule-based
approach. \newcite{Chen2010} build on top of their work to develop a
machine learning method. \newcite{Li2014text} implement a rule-based
system to detect the stimuli in Weibo posts and further inform an
emotion classifier with the output of this system. Other approaches to
develop rules include manual strategies \cite{Gao2015}, bootstrapping
\cite{Yada2017} and the use of constituency and dependency parsing
\cite{Neviarouskaya2013}.

All recently published state-of-the-art methods for the task of
emotion stimulus detection via clause classification are evaluated on
the Mandarin data by \newcite{Gui2016}. They include multi-kernel
learning \cite{Gui2016} and long short-term memory networks (LSTM)
\cite{Cheng2017}. \newcite{Gui2017} propose a convolutional
multiple-slot deep memory network (ConvMS-Memnet), and
\newcite{Li2018co} a co-attention neural network model, which encodes
the clauses with a co-attention based bi-directional long short-term
memory into high-level input representations, which are further passed
into a convolutional layer.  \newcite{Ding2019} proposed an
architecture with components for ``position augmented embedding'' and
``dynamic global label'' which takes the relative position of the
stimuli to the emotion keywords and use the predictions of previous
clauses as features for predicting subsequent clauses.
\newcite{Xia2019b} integrate the relative position of stimuli and
evaluate a transformer-based model that classifies all clauses jointly
within a text.  Similarly, \newcite{Yu2019} proposes a
word-phrase-clause hierarchical network.  The transformer-based model
achieves state of the art, however, it is shown that the RNN based
encoders are very close in performance \cite{Xia2019b}. Therefore, we
use a comparable model that is grounded on the same concept of a
hierarchical setup with LSTMs as encoders.  Further, there is a strand
of research which jointly predicts the clause that contains the
emotion stimulus together with its emotion cue
\cite{wei2020,fan2020}. However, the comparability of methods across
data sets has been limited in previous work, as
Table~\ref{tab:comparison} in the appendices shows.

\section{Conclusion} \label{sec:conclusions}
We contributed to emotion stimulus detection in two ways.  Firstly, we
evaluated emotion stimulus detection across several English annotated
data sets. Secondly, we analyzed if the current standard formulation
for stimulus detection on Mandarin Chinese is also a good choice for
English.

We find that the domain and annotation of the data sets have a large
impact on the performance. The worst performance of the token sequence
labeling approach is obtained on the crowdsourced data set \dsET. The
well-formed sentences of \dsES pose fewer difficulties to our models
than tweets and headlines.  We see that the sequence labeling
approaches are more appropriate for the phenomenon of stimulus
mentions in English. This shows in the evaluation of the comparably
coarse-grained clause level and is also backed by our error analysis.

For future work, we propose closer investigation of whether other
smaller constituents might represent the stimulus better for English
and a check of whether the strong results for the sequence labeling
hold for other languages. Notably, the clause classification setup has
its benefits, and this might lead to a promising setting as joint
modeling or as a filtering step to finding parts of the text which
might contain a stimulus mention. Another step is to investigate if
the emotion stimulus and the emotion category classification benefit
from joint modeling in English as it has been shown for Mandarin
\cite{Chen2018}.

\section*{Acknowledgments}
This research has been conducted within the project SEAT (Structured
Multi-Domain Emotion Analysis from Text, KL 2869/1-1), funded by the
German Research Council (DFG). We thank Enrica Troiano, Evgeny Kim,
Gabriella Lapesa, and Sean Papay for fruitful discussions and feedback
on earlier versions of the paper.

\appendix

\begin{figure*}
\section{Appendix}
\label{sec:appendix}
\centering
\small
\setlength{\tabcolsep}{3.5pt}
\renewcommand{\arraystretch}{0.95}
\begin{tabular}{llrrrrrrrrrrrrrrr}
\toprule
& & \multicolumn{12}{c}{SL Evaluation}&\multicolumn{3}{c}{CC Evaluation} \\
\cmidrule(lr){3-14}\cmidrule(lr){15-17}
&&\multicolumn{3}{c}{Exact}&\multicolumn{3}{c}{Relaxed}&\multicolumn{3}{c}{Left-Exact}&\multicolumn{3}{c}{Right-Exact}&\multicolumn{3}{c}{Clause}  \\
\cmidrule(r){1-1}\cmidrule(lr){2-2}\cmidrule(lr){3-5}\cmidrule(lr){6-8}\cmidrule(lr){9-11}\cmidrule(lr){12-14}\cmidrule(lr){15-17}
Data  & Model & P & R & \F & P & R & \F & P & R & \F & P & R & \F & P & R & \F \\
\cmidrule(r){1-1}\cmidrule(lr){2-2}\cmidrule(lr){3-5}\cmidrule(lr){6-8}\cmidrule(lr){9-11}\cmidrule(lr){12-14}\cmidrule(lr){15-17}
\multirow{3}{*}{\dsES}
 & SL   & 69 & 73 & 71 & 69 & 74 & 72 & 100 & 74 & 85 & 100 & 59 & 74 & 81 & 83 & 82  \\
 & ICC  & 03 & 26 & 05 & 10 & 100 & 18 & 03 & 26 & 05 & 05 & 44 & 09 & 82  & 70 & 73  \\
 & JCC  & 05 & 12 & 07 & 21 & 42 & 30 & 12 & 10 & 12 & 17 & 46 & 24 & 84 & 80 & 82 \\
\midrule
\multirow{3}{*}{\dsET}
& SL   & 15 & 07 & 10 & 41 & 30 & 35 & 52 & 09 & 15 & 42 & 11 & 17 & 100 & 40 & 57  \\
& ICC  & 12 & 47 & 19 & 22 & 100 & 37 & 13 & 47 & 21 & 21 & 74 & 32 & 59 & 59 & 59  \\
& JCC  & 14 & 30  & 19 & 25 & 54 & 34 & 15 & 48 & 23 & 28 & 47 & 35 & 59 & 57 & 58 \\
\midrule
\multirow{3}{*}{\dsECA}
& SL   & 16 & 20 & 18 & 42 & 60 & 49 & 76 & 29 & 41 & 83 & 23 & 36 & 99 & 68 & 80  \\
& ICC  & 05 & 35 & 09 & 14 & 100 & 24 & 05 & 35 & 09 & 13 & 82 & 22 & 79 & 64 & 67  \\
& JCC  & 06 & 29 & 10 & 18 & 40 & 25 & 11 & 15 & 12 & 35 & 29 & 31 & 82 & 68 & 74  \\
\midrule
\multirow{3}{*}{\dsGNE}
& SL  & 39 & 30 & 34 & 66 & 92 & 77 & 79 & 30 & 44 & 86 & 86 & 86 & 96 & 99 & 97  \\
& ICC & 15 & 29 & 19 & 37 & 100 & 54 & 15 & 29 & 20 & 48 & 92 & 63 & 71 & 67 & 68  \\
& JCC & 16 & 25 & 19 & 40 & 90 & 55 & 17 & 25 & 20 & 54 & 82 & 65 & 82 & 93 & 87  \\
\bottomrule
\end{tabular}
\caption{Results of the three different models across the
  five different datasets}
\label{tab:exp1}
\end{figure*}

\begin{figure*}
\centering\small
\newcommand{\x}{+}
\newcommand{\m}{\textcolor{lightgray}{$-$}}
\begin{tabular}{l|llcccccc}
  \toprule
  \multicolumn{1}{c}{}& & & \multicolumn{6}{c}{Data sets and Annotation Approach } \\
  \cmidrule(lr){4-9}
  \multicolumn{1}{c}{}& & & Categorical Class. &&&&&\\
  \multicolumn{1}{c}{}& & & \& Sequence Lab. & \multicolumn{3}{c}{Sequence Labeling} & \multicolumn{2}{c}{Clause Class.}\\
  \cmidrule(lr){4-4}\cmidrule(lr){5-7}\cmidrule(lr){8-9}
  \multicolumn{1}{c}{}& & & ET & ES & REMAN & GNE & ECA & EDCE \\
  \multicolumn{1}{c}{}& Models & Papers & (en) & (en) & (en) & (en) & (en) & (zh) \\
  \cmidrule(lr){2-2}\cmidrule(lr){3-3}\cmidrule(lr){4-4}\cmidrule(lr){5-5}\cmidrule(lr){6-6}\cmidrule(lr){7-7}\cmidrule(lr){8-8}\cmidrule(lr){9-9}
  \multirow{15}{*}{\rtb{Methods}}
                      & CRF & \newcite{Ghazi2015} & \m & \x & \m & \m & \m & \m \\
                      & BiLSTM-CRF & \newcite{Kim2018} & \m & \m & \x & \m  & \m & \m  \\
                      & BiLSTM-CRF & \newcite{Bostan2020} & \m & \m & \m & \x & \m & \m  \\
                      & SVM & \newcite{Mohammad2014} & \x & \m & \m & \m  & \m & \m  \\
                      & CRF & \newcite{Gao2017} & \m & \m & \m & \m & \x & \m  \\
                      & LSTM & \newcite{Cheng2017} & \m & \m & \m & \m & \m & \m  \\
                      & JMECause & \newcite{Chen2018} & \m & \m & \m & \m & \m & \m \\
                      & multi-kernel SVM & \newcite{Xu2017} & \m & \m & \m & \m & \m & \x  \\
                      & Multi-Kernel & \newcite{Gui2016} & \m & \m & \m & \m & \m & \x \\
                      & ConvMS-Memnet & \newcite{Gui2017} & \m & \m & \m & \m & \m & \x \\
                      & CANN & \newcite{Li2018co} & \m & \m & \m & \m & \m & \x \\
                      & PAE-DGL & \newcite{Ding2019} & \m & \m & \m & \m & \m & \x \\
                      & HCS & \newcite{Yu2019} & \m & \m & \m & \m & \m & \x \\
                      & Ranking & \newcite{Xu2019} & \m & \m & \m & \m & \m & \x \\
                      & Hierarchical BiLSTM & \newcite{Xia2019a} & \m & \m & \m & \m  & \m & \x \\
                      & RTHN & \newcite{Xia2019b} & \m & \m & \m & \m & \m & \x \\
                      & \textbf{Our work} & \textbf{Ours (2020)} & \textbf{\x} & \textbf{\x} & \textbf{\m} & \textbf{\x} & \textbf{\x} & \textbf{\m}  \\
                      & TransECPE & \newcite{fan2020} & \m & \m & \m & \m & \m & \x \\
                      & RankCP & \newcite{wei2020} & \m & \m & \m & \m & \m & \x
  \\
  \bottomrule
\end{tabular}
\caption{Mapping of previous state-of-the-art methods to data
  sets. \x\ indicates that we are aware of a publication which
  reports on the method being evaluated on the respective data set
  and a \m\ indicates our assumption that no reported results
  exist with the respective method being evaluated on the
  respective data set. ET corresponds to \dsET, ES to \dsES, GNE to
  \dsGNE, whereas the other data set are as being introduced
  above.}
\label{tab:comparison}
\end{figure*}

\end{document}